%% file: iclr2025_conference.tex
\documentclass{article} 
\usepackage{iclr2025_conference,times}

\input{math_commands.tex}

\usepackage{hyperref}
\usepackage{url}
\usepackage{booktabs}
\usepackage{multirow}
\usepackage{graphicx}

\title{OmniActor: A Generalist GUI and Embodied Agent for 2D\&3D Worlds}

\author{Longrong Yang\textsuperscript{1,2},
Zhixiong Zeng\textsuperscript{1}$^{*}$,
Yufeng Zhong\textsuperscript{1},
Jing Huang\textsuperscript{1},
Liming Zheng\textsuperscript{1}, 
\\
\textbf{Lei Chen\textsuperscript{1}}\textbf{,}
\textbf{Haibo Qiu\textsuperscript{1}}\textbf{,}
\textbf{Zequn Qin\textsuperscript{2}}\textbf{,}
\textbf{Lin Ma\textsuperscript{1}}$^{\dagger}$\textbf{,}
\textbf{Xi Li\textsuperscript{2}}$^{\dagger}$
\\
\textsuperscript{1} Meituan, \textsuperscript{2} Zhejiang University
}

\iclrfinalcopy 
\begin{document}

\footnotetext{$^{\dagger}$Corresponding authors. $^*$Project leader.}

\maketitle

\begin{abstract}
Multimodal large language models are evolving toward multimodal agents capable of proactively executing tasks. Most agent research focuses on GUI or embodied scenarios, which correspond to agents interacting with 2D virtual worlds or 3D real worlds, respectively. However, many complex tasks typically require agents to interleavely interact with these two types of environment.
We initially mix GUI and embodied data to train, but find the performance degeneration brought by the data conflict.
Further analysis reveals that GUI and embodied data exhibit synergy and conflict at the shallow and deep layers, respectively, which resembles the cerebrum-cerebellum mechanism in the human brain.
To this end, we propose a high-performance generalist agent \textbf{OmniActor}, designed from both structural and data perspectives.
First, we propose Layer-heterogeneity MoE to eliminate the conflict between GUI and embodied data by separating deep-layer parameters, while leverage their synergy by sharing shallow-layer parameters. 
By successfully leveraging the synergy and eliminating the conflict, OmniActor outperforms agents only trained by GUI or embodied data in GUI or embodied tasks. 
Furthermore, we unify the action spaces of GUI and embodied tasks, and collect large-scale GUI and embodied data from various sources for training. 
This significantly improves OmniActor under different scenarios, especially in GUI tasks. 
The code will be publicly available.
\end{abstract}

\input{sec/1_intro}
\input{sec/2_related}

\input{sec/3_methodology}
\input{sec/4_experiment}
\input{sec/5_conclusion}

\bibliography{iclr2025_conference}
\bibliographystyle{iclr2025_conference}

\end{document}

%% file: math_commands.tex
\usepackage{amsmath,amsfonts,bm}

\def\eqref#1{equation~\ref{#1}}

\def\1{\bm{1}}

\DeclareMathAlphabet{\mathsfit}{\encodingdefault}{\sfdefault}{m}{sl}
\SetMathAlphabet{\mathsfit}{bold}{\encodingdefault}{\sfdefault}{bx}{n}



%% file: sec/1_intro.tex
\section{Introduction}
Foundation models have demonstrated strong abilities in language and image understanding tasks. 
In particular, Multimodal Large Language Models (MLLMs)~\citep{liu2023visual,liu2024improved,wang2024qwen2,bai2025qwen2}, \textit{i.e.}, multimodal foundation models trained on massive amounts of image-text data, have shown excellent performance in image-text understanding tasks. 
As an extension of MLLMs, multimodal agents have been successfully applied to gragh user interface (GUI) control~\citep{wuatlas,xu2025aguvis,qin2025ui} and the field of embodied intelligence~\citep{shahmutex,ha2023scaling,kim2025openvla}, and have begun to be widely deployed in industry.

Humans can perform actions in different environments, such as completing shopping in a 2D digital world or interacting with entities in a 3D physical world. 
Therefore, a natural question arises: \textit{can an agent perform actions in both 2D and 3D worlds like humans, thereby achieving a generalist agent?}
This is challenging because there are inherent differences between the 2D world and the 3D world.
GUI and embodied tasks are two typical tasks for evaluating agent abilities in 2D and 3D worlds, respectively.
Some works have made preliminary explorations on unifying GUI and embodied action abilities into an agent. 
For instance, Magma~\citep{yang2025magma} unifies GUI and embodied data into Set-of-Mark (SoM) and Trace-of-Mark (ToM), and uses videos to augment action data. 
GEA~\citep{szot2025multimodal} employs a continuous multi-embodiment tokenizer for mixed action prediction, and sets online reinforcement learning stage to improve the agent.

\begin{figure}[tb]
\begin{center}
\includegraphics[width=0.9\linewidth]{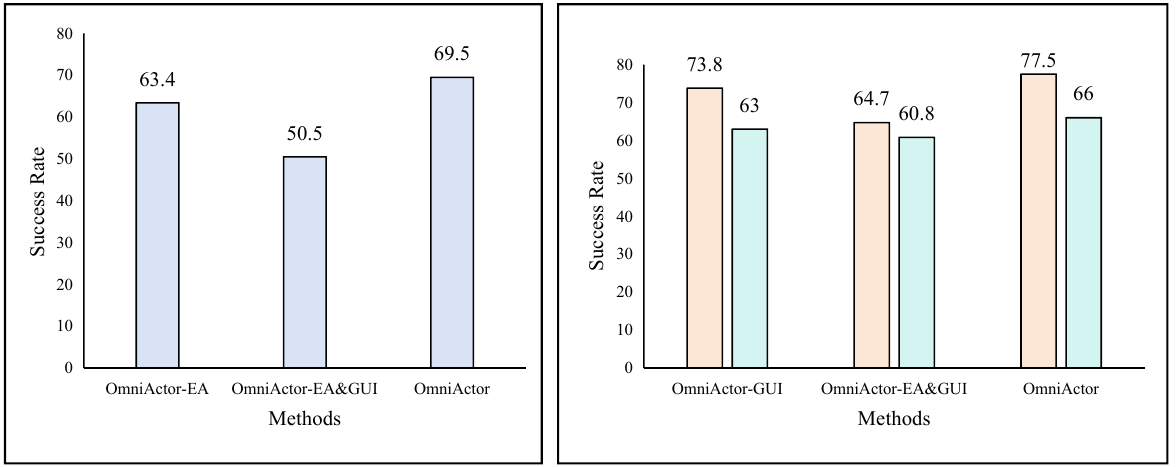}
\end{center}
  \caption{Performance analysis.
  OmniActor-GUI means the agent trained on GUI data, building on Qwen2-VL.
  OmniActor-EA means the agent trained on embodied data.
  OmniActor-EA\&GUI means the agent trained on GUI and embodied data.
  OmniActor uses the proposed Layer-heterogeneity MoE to leverage synergy and eliminate conflict, significantly improving the agent.
  }
\label{teaser}
\end{figure}

When unifying data from different environments into one model, a fundamental issue is the conflict and synergy among these data.
We initially mix GUI and embodied data to train, but find the performance degeneration brought by the data conflict, as shown in Figure~\ref{teaser}.
We attribute the conflict to the significant action differences between GUI and embodied tasks.
In specific, GUI actions usually consist of a series of text-described operations (\textit{e.g.}, "click"), while embodied actions usually involve a sequence of 6-DoF displacements of the end effector. 
On the other hand, GUI and embodied tasks may exhibit synergy, as they share a similar task structure (environments, instructions, and actions).
The understanding of environments and instructions may mutually promote each other in embodied and GUI tasks.
However, existing generalist agents lack specific mechanisms to handle conflict and synergy, leading to the suboptimal performance for both tasks.

Furthermore, we find that single-environment (\textit{i.e.}, GUI or embodied) data used by existing generalist agents are insufficient. Existing generalist agents focus on expanding the action environments of agents. 
However, training data used in these methods in terms of GUI or embodied actions is significantly lower than that of single-environment agents, \textit{e.g.}, Aguvis~\citep{xu2025aguvis}. 
This makes that their performance is significantly lower than state-of-the-art single-environment agents in GUI or embodied tasks, making it difficult to deploy them in industry.

To this end, we propose a novel generalist agent, naming OmniActor, aiming to utilize the synergy between GUI and embodied data while eliminating their conflict.
Specifically,
$(i)$ \textbf{Layer-heterogeneity MoE}. 
By investigating the parameter update directions from different data, we observe that compared to deep layers, the parameter update directions from GUI and embodied data in shallow layers are more consistent. 
This can be analogous to the cerebrum-cerebellum mechanism for humans: the cerebrum, which is closer to the input, performs comprehensive understanding of environments and instructions. 
In contrast, the cerebellum, which is closer to the output, needs to execute different actions.
Thus, we share parameters in shallow layers to leverage the synergy between GUI and embodied data, and separate parameters in deep layers to eliminate their conflicts caused by action differences.
This significantly improves the agent performance, as shown in Figure~\ref{teaser}.
$(ii)$ \textbf{Large-scale GUI and embodied data}. 
We first collect data from various sources, including GUI data sources OS-Atlas~\cite{wuatlas}, Uground~\cite{gou2025uground}, Aguvis~\cite{xu2025aguvis}, and Aria-UI~\cite{yang-etal-2025-aria} and embodied data source LIBERO~\cite{liu2023libero}. 
All data are generalist into the same format, where each sample includes a system prompt, an image (environment), a task instruction, and an output action. 
We use the text tokenizer and the special embodied tokenizer to convert GUI actions and embodied actions into tokens in the same vocabulary, respectively.
By unifying data formats and action spaces, we use large-scale GUI and embodied data to train the agent, enhancing its ability to perform tasks.

In summary, our contributions are as follows:
\begin{itemize}
\item We propose the Layer-heterogeneity MoE. The Layer-heterogeneity MoE shares parameters in the shallow layers to leverage the synergy between GUI and embodied data, while separating parameters in the deep layers to eliminate conflicts between GUI and embodied data caused by action differences.

\item We unify and expand GUI and embodied data. By unifying data formats and action spaces, we can train the agent with large-scale GUI and embodied data, thereby significantly improving the performance of agents on GUI and embodied tasks.

\item We extend agent abilities on MLLMs to implement a generalist agent OmniActor. OmniActor significantly outperforms existing generalist agents in GUI and embodied tasks, and even shows the superiority than state-of-the-art single-environment agents.
\end{itemize}

%% file: sec/2_related.tex
\section{Related Works}
\subsection{GUI agent}
Early GUI agents~\citep{deng2023mind2web,gur2024real,lai2024autowebglm,he2024webvoyager,yang2023setofmark} rely on HTML or AXTree data to describe GUI elements through textual descriptions.
These methods depended on the structured representation of web page elements, enabling agents to locate targets based on tags, attributes, or text content.
However, these methods need meticulous pre-processing designs, limiting their generalization under different scenarios.
With the rise of multimodal large language models (MLLMs), an increasing number of pure vision-based methods have been proposed.
Leveraging the strong visual capabilities of MLLM, these methods eliminate the need for manually designing data pre-processing schemes for each scenario, and thus have significant advantages over HTML-dependent GUI agents in terms of generalization.
Early pure visual GUI agents~\citep{hong2024cogagent,zhang2024you,zhang2024android} focus on using MLLMs to process GUI screenshots to understand GUI components, replacing HTML-dependent GUI component understanding. 
For example, CogAgent~\citep{hong2024cogagent} uses MLLMs to process high-resolution GUI screenshots, achieving performance that surpasses HTML-based methods. Auto-GUI~\citep{zhang2024you} unifies GUI grounding into a text-driven grounding task and proposes action chains, using a series of historical actions to assist agents in decision-making.
Furthermore, researchers find that scaling data is crucial for enhancing GUI agents, thus proposing the use of synthetic data or video data~\citep{yang-etal-2025-aria,xie2025gui,wuatlas,sun2025gui}. 
For instance, Aria-UI~\citep{yang-etal-2025-aria} introduces an extensible data synthesis pipeline for generating grounding instruction samples, which are used to train MLLMs specifically for GUI grounding. OS-Atlas~\citep{wuatlas} releases the first multi-platform GUI data synthesis toolkit, supporting the automatic synthesis of cross-platform GUI grounding data while resolving action naming conflicts. GUI-Xplore~\citep{sun2025gui} enables GUI agents to learn from videos.

Recently, researchers have aimed to make GUI agents more similar to human-like agents~\citep{gou2025uground,xu2025aguvis,lu2025gui,qin2025ui}. 
For example, UGround~\citep{gou2025uground} proposes to execute actions only using human-like keyboard and mouse operations, demonstrating the feasibility of human-like GUI agents. Aguvis~\citep{xu2025aguvis} unifies different GUI action spaces and divides training into grounding and planning, enhancing the planning capabilities of agents after they have acquired strong GUI grounding abilities. GUI-Odyssey~\citep{lu2025gui} proposes a large-scale cross-application training and evaluation dataset, allowing agents to interleavely use multiple applications to perform tasks instead of being limited to a single application. UI-TARS~\citep{qin2025ui} can perform human-like interactions, which achieves accurate perception of GUI elements by collecting a large amount of GUI screenshots and enhances agent capabilities through multiple reasoning modes.

\subsection{Embodied agent}
In the field of robotics, benefiting from the rapid development of MLLMs, there has emerged a new trend of leveraging the strong generalization ability of MLLMs to enable them to act as ``brains" for controlling robots to perform tasks.
ATM~\citep{wen2023atm} uses pre-trained models to predict trajectories in videos, providing action guidance for agents.
MUTEX~\citep{shahmutex} learns strategies from multiple modalities, which can understand instructions in six modalities or their combinations.
ACT~\citep{zhaolearning} proposes a low-cost system that enables robots to complete multiple high-difficulty tasks through Supervised Fine-Tuning (SFT) and a Transformer-based action chunking algorithm.
Distill-D~\citep{ha2023scaling} efficiently generates embodied data with language labels using Large Language Models (LLMs) and sampling planners.
MaIL~\citep{jia2024mail} proposes a new SFT architecture based on Mamba, which reduces overfitting and enhance generalization.
Prise~\citep{zheng2024prise} regards temporal action abstraction as a sequence compression problem, and combines continuous action quantization with byte pair encoding to learn action abstractions.
MDT~\citep{reuss2024multimodal} addresses data annotation issues through potential goal-conditional state representations and can handle long-term manipulation tasks with sparsely annotated data.
Then, RoboFlamingo~\citep{li2024vision}, Octo~\citep{team2024octo}, and OpenVLA~\citep{kim2025openvla} are vision-language-action models built on open-source MLLMs, which can be applied to robot control through fine-tuning.
GR-2~\citep{cheang2024gr} pre-trains agents on massive internet video data and fine-tune agents on embodied trajectory data.

\subsection{Generalist Agent}
Researchers have recently begun to study generalist agents, aiming to develop a foundational model that serves multimodal agent tasks in both digital and physical worlds, capable of accomplishing various agent tasks ranging from GUI navigation to robotic manipulation.
Magma~\citep{yang2025magma} unifies GUI and embodied tasks into State of Motion (SoM) and Trajectory of Motion (ToM), and augments action trajectory data with video data.
GEA~\citep{szot2025multimodal} utilizes a Continuous Multi-Embodiment Tokenizer to perform hybrid action prediction for embodied intelligence, games, GUI control, planning, and navigation, and sets Supervised Fine-Tuning and Online Reinforcement Learning.
NaviMaster~\citep{luo2025navimaster} holds that both GUI and embodied tasks can be formulated as Markov decision processes, thus enabling the generation of trajectories for GUI and embodied tasks in a generalist form. Based on this, it employs reinforcement learning and a distance-aware reward to enhance the generalization of the agent.

%% file: sec/3_methodology.tex
\section{Methodology}

\subsection{Overview}
\textbf{generalist Agent}:
A multimodal generalist agent should be able to interact with both 2D virtual worlds and 3D real worlds.
We define a multimodal agent ${\pi}$, whose inputs are the visual observation $I$ for the environment, as well as task descriptions $T$ and historical information $H$ in text form, and its output is a set of actions $A$, specifically as follows:
\begin{equation}
    A = \mathcal{\pi}(I, T, H) = \{a^{l}_{1},\cdot\cdot\cdot,a^{l}_{N}\},
\end{equation}
where $l=\{\text{gui},\text{robot}\}$ indicates whether the $n$-th action token $a_{n}$ is a GUI token or a embodied token.
This formula is applicable to different tasks:

\begin{itemize}
\item GUI tasks in 2D virtual worlds: The GUI task $T$ may be ``book a hotel", and the output includes both language tokens representing the semantic type of the action (\textit{e.g.}, ``tap" or ``click") and the position (x, y) where the action is applied.

\item Embodied tasks in 3D real worlds: The embodied (robot control) task $T$ may be ``put the black bowl at the back on the plate". The output includes a 6-degree-of-freedom (6-DoF) displacement $(\text{pos}_x,\text{pos}_y,\text{pos}_z,\text{rot}_x,\text{rot}_y,\text{rot}_z)$ of the end effector and an additional dimension to indicate whether the gripper is open or closed, where $(\text{pos}_x,\text{pos}_y,\text{pos}_z)$ indicates the spatial position of the robotic arm and $(\text{rot}_x,\text{rot}_y,\text{rot}_z)$ indicates the rotation angle of the robotic arm.
Each dimension usually ranges from [-1,1] after normalization.
\end{itemize}
Given images (observation), task instruction, and history information, a multimodal generalist agent should predict GUI actions or embodied actions based on environments.
Thus, it is important to generalist different actions into a action space, which will be specified below.

\begin{figure}[tb]
\begin{center}
\includegraphics[width=1.0\linewidth]{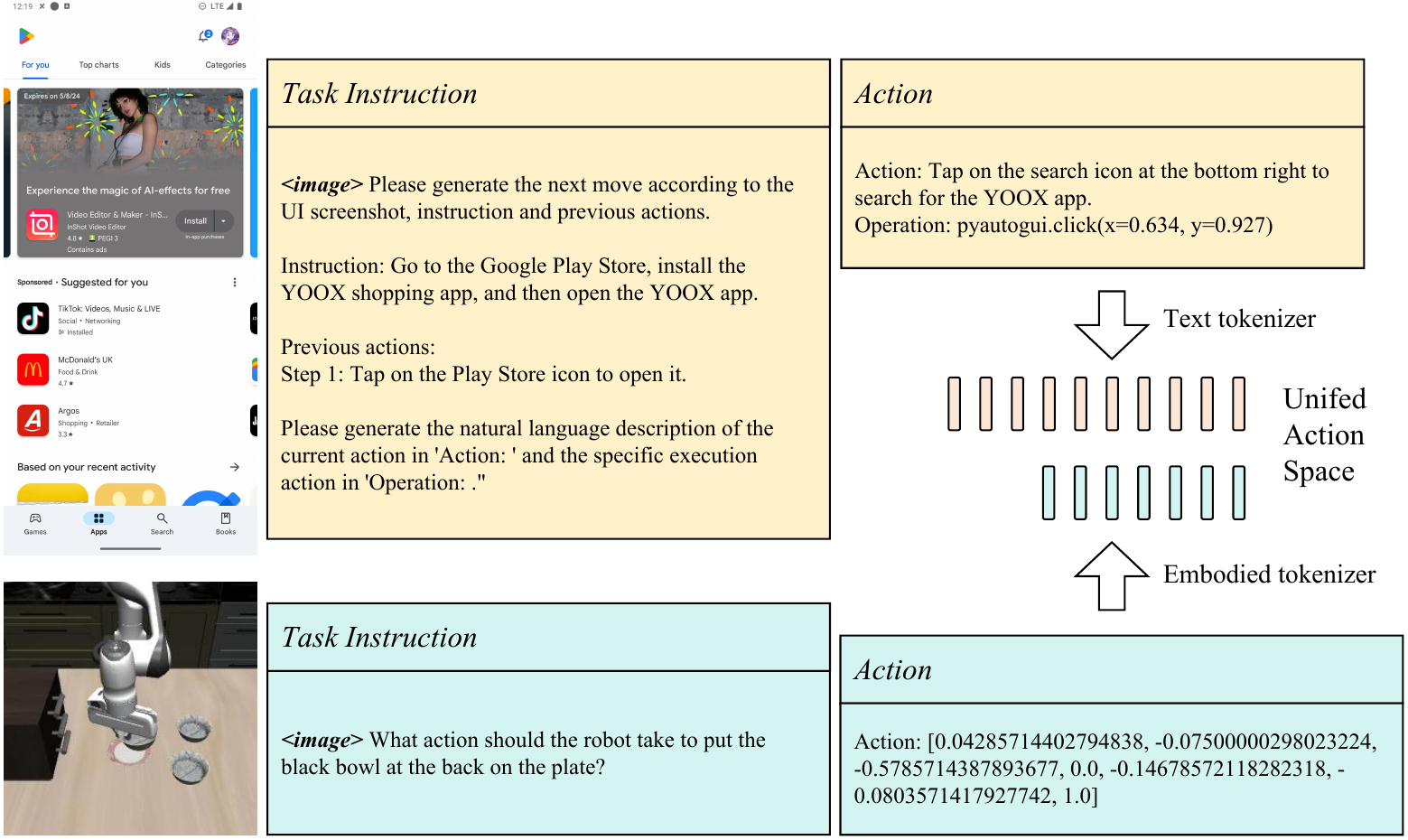}
\end{center}
  \caption{Data format.
  We unify GUI tasks in 2D digital world and embodied tasks in 3D real world.
  Each sample includes a system prompt, an image (environment), a task instruction, and an action description.
  GUI actions are usually described by texts.
  We directly use the tokenizer of MLLM as GUI tokenizer, converting actions into tokens.
  Embodied actions are usually described by 6-DoF displacement of the end effector and an additional gripper control signal.
  We use embodied tokenizer to convert actions into tokens.
  GUI and embodied actions share the same vocabulary, thus being in a generalist action space for predicting.
  }
\label{format}
\end{figure}

\begin{figure}[tb]
\begin{center}
\includegraphics[width=1.0\linewidth]{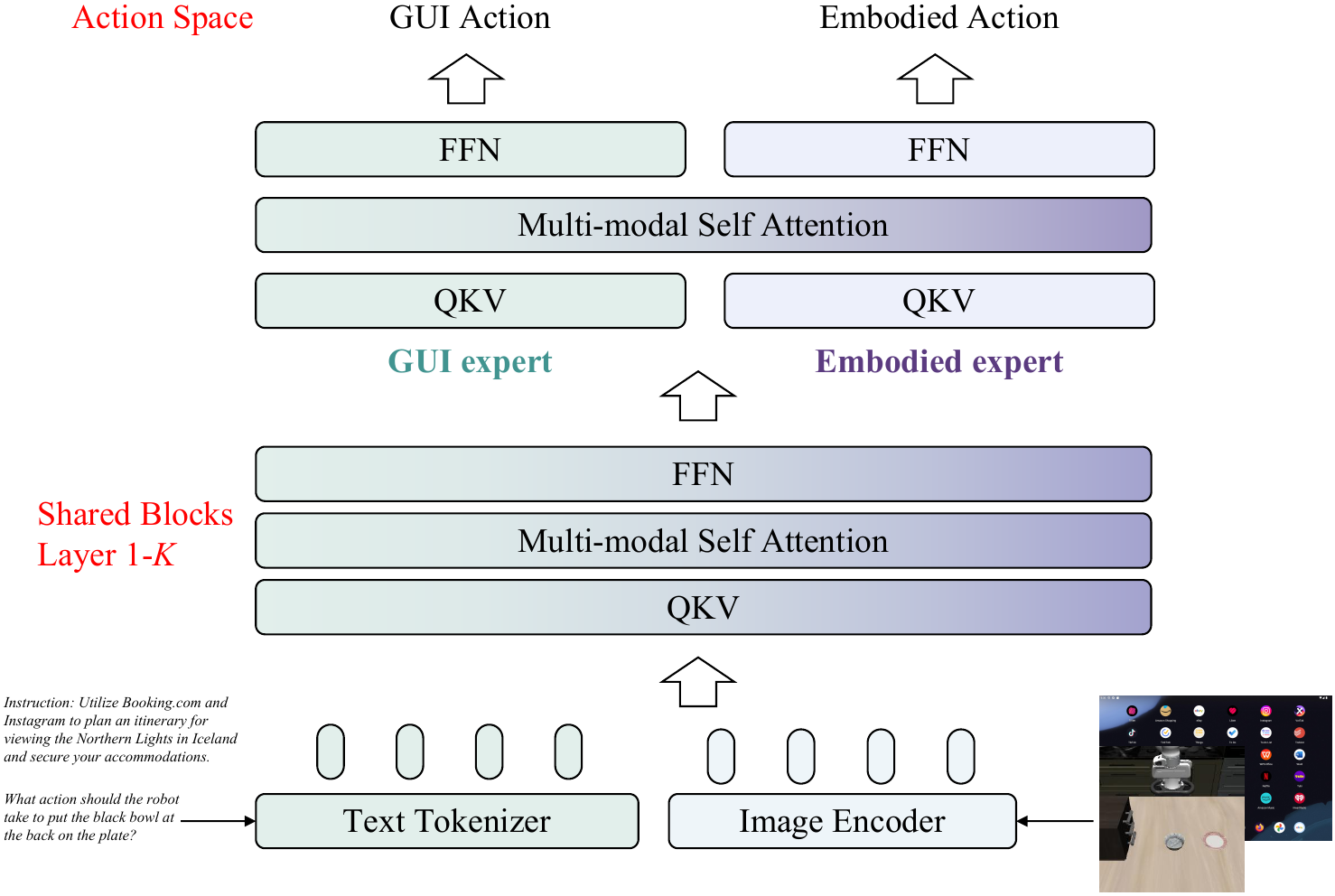}
\end{center}
  \caption{The proposed pipeline.
  The proposed \textbf{Layer-heterogeneity MoE} shares parameters in shallow layers to leverage the synergy between GUI and embodied data, and separates parameters in deep layers to eliminate conflicts between GUI and embodied data caused by action differences.
  }
\label{pipeline}
\end{figure}

\textbf{Our Method}: To train a generalist agent OmniActor, while leveraging the synergy between GUI and embodiment and eliminating their conflicts, we have made improvements in both data and structure:
$(i)$ Data. As shown in Figure~\ref{format}, we unify GUI and embodied data into the same format, where each sample includes a system prompt, an image, a task instruction, and an action description. 
Embodied actions are represented by values in [-1, 1], and we uniformly discretize [-1, 1] into K intervals, with each interval corresponding to a token ID, thereby converting embodied actions into tokens.
$(ii)$ Structure. We propose the layer-heterogeneity MoE. 
As shown in Figure~\ref{pipeline}, the layer-heterogeneity MoE shares shallow-layer parameters to utilize the synergy between GUI and embodied data, while separating deep-layer parameters to eliminate conflicts between GUI and embodied data caused by action differences. 
This structure is similar to the cerebrum and the cerebellum: the part close to the input side, like the cerebrum, comprehensively processes information, and the part close to the action side, like the cerebellum, generates different actions respectively.

The details of these modules will be introduced in detail in the subsequent sections.

\subsection{Data Processing}

\textbf{Unification of Data Format}: Qwen2-VL~\citep{wang2024qwen2} and Qwen2.5-VL~\citep{bai2025qwen2} are powerful and widely used MLLMs, and we adopt them as our base MLLMs. 
To be compatible with Qwen2-VL or Qwen2.5-VL, we convert all GUI and embodied data into the ShareGPT format, as shown in Figure~\ref{format}.
Each sample includes a system prompt, an image, a task instruction, and an action. 
The system prompt represents the action space used by the agent, and the image represents the current environment where the agent is located.
Images are processed into tokens using the image encoder of the base MLLMs. 
System prompts, task instructions, and GUI actions are processed into tokens using the text tokenizer of the base MLLMs. Embodied actions undergo special processing.

\textbf{Tokenization of Embodied Actions}: We convert embodied actions into tokens for handling. 
An embodied action is a 6-DoF displacement $(\text{pos}_x,\text{pos}_y,\text{pos}_z,\text{rot}_x,\text{rot}_y,\text{rot}_z)$ of the end effector and an additional dimension to indicate whether the gripper is open or closed, where each dimension ranges from [-1, 1] after normalization.
We uniformly discretize [-1, 1] into K intervals. 
Then, we select the K token IDs with the lowest frequency from the vocabulary of the base MLLM, which correspond to the K intervals. 
For example, as shown in Figure~\ref{format}, if an embodied action is [0.043, -0.075, -0.579, 0.0, -0.147, -0.080, 1.0], the tokens obtained after conversion will be [151510, 151500, 151482, 151515, 151515, 151516, 151642].

Through the above two steps, we have prepared the GUI and embodied data, which can be used for training the base MLLM to produce a generalist agent.

\subsection{Layer-Heterogeneity MoE}
GUI and embodied data exhibit synergy and conflict, which, from the perspective of model optimization, are manifested in the parameter optimization directions. 
In specific, ``synergy" indicates the similar parameter optimization directions from GUI and embodied data, while ``conflict" indicates the different parameter optimization directions from GUI and embodied data. 
From this perspective, if GUI and embodiment are synergistic for some parameters, these parameters should be shared; otherwise, these parameters should be separated. 
Based on the basic idea, we analyze parameter optimization directions between GUI and embodiment, and find that the optimization direction similarity in the shallow layers is much higher than that in the deep layers. 
Our hypothesis for this phenomenon is: the understanding of environments (images) and instructions is synergistic for GUI and embodied tasks. However, there is a significant difference between GUI actions and embodied actions on the output side, which requires processing with different parameters. 
Based on this phenomenon, we propose the layer-heterogeneity MoE, aiming to leverage the synergy between GUI and embodied data while eliminating their conflicts, as shown in Figure~\ref{pipeline}.

Specifically, assuming the layer count of the LLM is $L$, we set a layer depth threshold $K$ to distinguish shared layers and separated layers.

$(i)$ For shallow layers $l_{s}$ ($1 \leq l_{s} \leq K$), as shown in Figure~\ref{pipeline}, we share parameters to process both GUI and embodied data, for utilizing the synergy between GUI and embodied data:
\begin{equation}
    \mathbf{x}_{\ell}^{\prime} = \mathrm{MSA}(\mathrm{LN}(\mathbf{x}_{\ell-1}))+\mathbf{x}_{\ell-1},  \ell \in \{1, \ldots, K\},
\label{ffn}
\end{equation}
\begin{equation}
    \mathbf{x}_{\ell} = \mathrm{FFN}(\mathrm{LN}(\mathbf{x^{\prime}}_{\ell}))+\mathbf{x^{\prime}}_{\ell},  \ell \in \{1, \ldots, K\},
\label{ffn2}
\end{equation}
where $\mathbf{x}_{\ell}$ indicates the hidden representation in the layer $\ell$, MSA indicates the Multi-head Self-Attention, FFN indicates the Feed-Forward Network, and LN indicates the Layer Normalization.

$(ii)$ For deep layers $l_{d}$ ($K < l_{d} \leq L$), as shown in Figure~\ref{pipeline}, we separate the attention and FFN. 
One set of parameters (\textit{e.g.}, $\text{FFN}_{gui}$) is used to process GUI data, and another set of parameters (\textit{e.g.}, $\text{FFN}_{rob}$) is used to process embodied data:
\begin{equation}
\mathbf{x}_\ell' = 
\begin{cases} 
\mathrm{MSA_{gui}}\left(\mathrm{LN_{gui}}(\mathbf{x}_{\ell-1})\right) + \mathbf{x}_{\ell-1}, & \text{if } \mathbf{x}_{\ell-1} \text{ is GUI data} \\
\mathrm{MSA_{rob}}\left(\mathrm{LN_{rob}}(\mathbf{x}_{\ell-1})\right) + \mathbf{x}_{\ell-1}, & \text{if } \mathbf{x}_{\ell-1} \text{ is embodied data}
\end{cases}
\quad \ell \in \{K+1, \ldots, L\}
\end{equation}
\begin{equation}
\mathbf{x}_\ell = 
\begin{cases} 
\mathrm{FFN_{gui}}\left(\mathrm{LN_{gui}}(\mathbf{x}_{\ell}')\right) + \mathbf{x}_{\ell}', & \text{if } \mathbf{x}_{\ell}' \text{ is GUI data} \\
\mathrm{FFN_{rob}}\left(\mathrm{LN_{rob}}(\mathbf{x}_{\ell}')\right) + \mathbf{x}_{\ell}', & \text{if } \mathbf{x}_{\ell}' \text{ is embodied data}
\end{cases}
\quad \ell \in \{K+1, \ldots, L\}
\end{equation}
where $\mathrm{MSA_{gui}}$ ($\mathrm{FFN_{gui}}$, $\mathrm{LN_{gui}}$) and $\mathrm{MSA_{rob}}$ ($\mathrm{FFN_{rob}}$, $\mathrm{LN_{rob}}$) mean expert parameters for GUI and embodied tasks, respectively.
In addition, as shown in Figure~\ref{pipeline}, we use different heads $h_{gui}$ and $h_{rob}$ to predict GUI actions and embodied actions, respectively.
Although GUI and embodied actions are generalist in a single vocabulary, their action spaces differ significantly. 
Therefore, separating the heads is important for eliminating conflicts between GUI and embodied actions.
During inference, we assume that the task type of the sample (GUI or embodied task) is known, and we select the appropriate branch to process the data based on the task type of the sample.

Through the layer-heterogeneity MoE design with shallow sharing and deep separation, we effectively utilize the synergy between GUI and embodied data, while eliminating conflicts between them, thereby realizing a high-performance generalist agent for 2D and 3D worlds.

%% file: sec/4_experiment.tex
\section{Experiments}

\subsection{Experimental Setup}
\textbf{Training data}: 
Due to the diversity and complexity of GUI components, we first train the MLLM using GUI grounding data (a combination of open-source datasets OS-Atlas~\citep{wuatlas}, Uground~\citep{gou2025uground}, Aguvis~\citep{xu2025aguvis}, and Aria-UI~\citep{yang-etal-2025-aria}, similar to ScaleTrack~\citep{huang2025scaletrack}). 
This stage is not set for embodied tasks because the components in embodied scenarios are relatively simple. 
After this stage, we use trajectory data to train the model, enabling it to become an agent for performing tasks. 
For GUI tasks, we follow ScaleTrack~\citep{huang2025scaletrack} to select Aguvis~\citep{xu2025aguvis}; for embodied tasks, we select the popular LIBERO dataset~\citep{liu2023libero}. 
We unify GUI and embodied data into the same format, and also unify the action spaces of GUI and embodied tasks, for training a generalist agent. 
During training, since embodied actions consist of a series of continuous points, which differ greatly from the output of MLLM, it is necessary to resample the embodied data to ensure sufficient learning of embodied tasks~\citep{zawalski2025robotic,szot2025multimodal}. 
We resample the embodied data 5 times, and the ratio of GUI data to embodied data is approximately 1:5.
The data size is about 3.4M for grounding training and about 4.1M for trajectory training.

\textbf{Training hyper-parameters}: 
We use the AdamW optimizer with a learning rate of 1e-5 and a cosine learning rate scheduler with a warm-up ratio of 0.03. 
We set the global batch size to 128 in the grounding training phase and 64 in the trajectory training phase, and adopt the DeepSpeed ZERO3-style parallel strategy.
To align with GUI data, we resize all 256$\times$256 images from LIBERO to 448$\times$448.
The layer depth threshold $K$ to distinguish shared and separated layers is set to 8.

\textbf{Evaluation benchmarks}:
We select two datasets as the GUI evaluation benchmark: $(i)$ Android Control~\citep{li2024effects}. 
Android Control is designed to evaluate GUI agents on mobile platforms. 
It consists of two types of sub-tasks: AndroidControl-Low requires the agent to predict specific action types and parameters (\textit{e.g.}, coordinates) at each step based on an image, a task instruction, and a human-annotated natural language description of actions.
AndroidControl-High requires the agent to independently plan and sequentially predict action types and parameters when only given images and task instructions. 
$(ii)$ GUI-Odyssey~\citep{lu2025gui} provides a comprehensive dataset for evaluating cross-application GUI agents. Similar to AndroidControl-High, it only provides images and task instructions, and it requires the agent to perform complex cross-application operations.
Following previous studies, we randomly sample 800-step actions to create a subset for evaluation.

We select LIBERO~\citep{liu2023libero} as the embodied evaluation benchmark, which includes a total of 90 evaluation tasks covering three different scenarios: kitchen, living room, and study room.
For each task, we test 20 unseen start and goal configurations. 
In each test, the agent first receives the task instruction and the initial scene (image), then outputs an action. The action and the scene are fed into the end effector to generates a new scene (image). The task instruction and the new scene are then fed into the agent for predicting the next action. When the task is successfully executed, the end effector returns a success signal, and the evaluation proceeds to the next task. The maximum steps allowed for task execution is 400.

\textbf{Comparison}:
We conduct a comprehensive comparison with GUI agents, embodied agents, and generalist agents.
$(i)$ In-house GUI agents, including Claude 3.5~\citep{hu2024dawn}, GPT-4o~\citep{hurst2024gpt}, Qwen2-VL-7B~\citep{wang2024qwen2}, and UI-TARS-7B~\citep{qin2025ui}.
Open-sourced GUI agents, including SeeClick~\citep{cheng2024seeclick}, Aria-UI~\citep{yang-etal-2025-aria}, OS-Atlas-7B~\citep{wuatlas}, Aguvis-7B~\citep{xu2025aguvis}, and ScaleTrack-7B~\citep{huang2025scaletrack}.

$(ii)$ Embodied agents, including MaIL~\citep{jia2024mail}, PRISE~\citep{zheng2024prise}, ATM~\citep{wen2023atm}, MUTEX~\citep{shahmutex}, ACT~\citep{zhaolearning}, Distill-D~\citep{ha2023scaling}, and MDT~\citep{reuss2024multimodal}. 

$(iii)$ generalist agents, including Magma~\citep{yang2025magma}, GEA~\citep{szot2025multimodal}, and NaviMaster~\citep{luo2025navimaster}.
In detail, Magma~\citep{yang2025magma} does not report results on LIBERO, AndroidControl, and GUI Odyssey, but provides codes and the pre-trained model.
Thus, we finetune the pre-trained model on LIBERO, AndroidControl, and GUI Odyssey to report the performance.
GEA and NaviMaster do not provide codes, so we only report the results in its paper.

\textbf{Our method}:
$(i)$ OmniActor-GUI: Use GUI data to train Qwen2-VL~\citep{wang2024qwen2}.
$(ii)$ OmniActor-EA: Use embodied data to train Qwen2-VL~\citep{wang2024qwen2}.
$(iii)$ OmniActor: Use GUI and embodied data to train Qwen2-VL~\citep{wang2024qwen2}.
Parameters in shallow layers are shared.
Parameters in deep layers are separated as the MoE, where GUI data is processed by one set of experts, and embodied data is processed by another set of experts.

\subsection{Main Results}

\begin{table*}[!t]
\small
\setlength\tabcolsep{1.5mm}
\caption{\textbf{Comparison between different agents on embodied (robot control) tasks and GUI navigation tasks.} 
We report the success rate on each task.
``-" indicates that the agent is unable to perform the task or its performance on the task has not been reported.
}
\label{tab:main}
\centering
\begin{tabular}{l|c|c|ccc} 
\toprule
\multirow{2}{*}{Agents} & \multirow{2}{*}{Source} & \textbf{Embodied Tasks} & \multicolumn{3}{c}{\textbf{GUI Tasks}} \\
& & LIBERO-90 & AndroidControl-Low & AndroidControl-High & GUI Odyssey\\
\midrule
Claude & In-house & - & 19.4 & 12.5 & 3.1 \\
GPT-4o & In-house & - & 19.4 & 20.8 & 3.3 \\
Qwen2-VL-7B & In-house & - & 82.6 & 69.7 & 60.2 \\
UI-TARS-7B & In-house & - & 90.8 & 72.5 & 87.0 \\
SeeClick & ACL'24 & - & 75.0 & 59.1 & 53.9 \\
Aria-UI & ACL'25 & - & 67.3 & 10.2 & 36.5 \\
OS-Atlas-7B & ICLR'25 & - & 85.2 & 71.2 & 62.0 \\
Aguvis-7B & ICML'25 & - & 80.5 & 61.5 & 63.8 \\
ScaleTrack-7B & arXiv'25 & - & 86.6 & 77.9 & 65.3 \\
\midrule
MUTEX & CoRL'23 & 53.0 & - & - & - \\
Distill-D & CoRL'23 & 49.9 & - & - & - \\
ACT & RSS'23 & 46.6 & - & - & - \\
MaIL & CoRL'24 & 60.3 & - & - & - \\
PRISE & ICML'24 & 54.4 & - & - & - \\
ATM & RSS'24 & 48.4 & - & - & - \\
MDT & RSS'24 & 67.2 & - & - & - \\
\midrule
Magma & CVPR'25 & 34.7 & 52.1 & 32.7 & 51.0 \\
GEA & CVPR'25 & 48.0 & - & 57.3 & - \\
NaviMaster & arXiv'25 & - & 68.9 & 54.0 & - \\
\midrule
OmniActor-GUI & - & - & 89.4 & 73.8 & 63.0 \\
OmniActor-EA & - & 63.4 & - & - & - \\
OmniActor & - & 69.5 & 86.4 & 77.5 & 66.0 \\
\bottomrule
\end{tabular}
\end{table*}

As shown in Table~\ref{tab:main}, we have several findings: 
$(i)$ Existing generalist agents usually have a relatively low performance than GUI agents on GUI tasks and embodied agents on embodied tasks, although they can finish the two tasks simultaneously.
$(ii)$ OmniActor-GUI achieves the comparable performance with existing GUI agents on GUI tasks.
OmniActor-EA achieves the comparable performance with existing embodied agents on embodied tasks.
$(iii)$ On embodied tasks, OmniActor achieves 6.1\% higher task success rate than OmniActor-EA.
On GUI tasks, OmniActor achieves the average 1.2\% higher task success rate than OmniActor-GUI.
Especially, OmniActor has a significantly higher task success rate on AndroidControl-High and GUI Odyssey, indicating the superiority on long-chain trajectory prediction.

\subsection{Study}

\begin{table*}[!t]
\small
\setlength\tabcolsep{1.5mm}
\caption{\textbf{Comparison between different parameter sharing and separation strategies.} 
We report the success rate on each task.}
\label{tab:study1}
\centering
\begin{tabular}{l|c|ccc|c} 
\toprule
\multirow{2}{*}{Models} & \textbf{Embodied Tasks} & \multicolumn{3}{c|}{\textbf{GUI Tasks}} & \multirow{2}{*}{Avg} \\
& LIBERO-90 & AndroidControl-Low & AndroidControl-High & GUI Odyssey & \\
\midrule
OmniActor-GUI & - & 89.4 & 73.8 & 63.0 & - \\
OmniActor-EA & 63.4 & - & - & - & - \\
\midrule
OmniActor-EA\&GUI & 50.5 & 85.3 & 64.7 & 60.8 & 65.3 \\
OmniActor \textit{hard} & 59.5 & 86.3 & 75.4 & 63.9 & 71.3 \\
OmniActor & 69.5 & 86.4 & 77.5 & 66.0 & 74.9 \\
\bottomrule
\end{tabular}
\end{table*}

\textbf{Study about parameter sharing and separation}:
We hope to utilize the synergy between GUI and embodied data and eliminate their conflicts by studying parameter sharing and separation.
We conduct the following experiments:
$(i)$ OmniActor-EA\&GUI: We directly mix GUI and embodied data to train Qwen2-VL~\citep{wang2024qwen2}.
$(ii)$ OmniActor \textit{hard}: Building on $(i)$, we modify the model structure to MoE, where GUI data is processed by one set of experts, and embodied data is processed by another set of experts.
The parameters of attention heads, FFNs, and classification head are totally separated.
$(iii)$ OmniActor: Different from $(ii)$, parameters in shallow layers are shared, and parameters in deep layers are separated as the MoE.

As shown in Table~\ref{tab:study1}, we have some important findings: 
$(i)$ Directly mixing GUI and embodied data to train models deteriorate the model performance, because GUI and embodied data are in conflict.
$(ii)$ OmniActor \textit{hard} separates almost all parameters for GUI and embodied tasks. 
Compared with direct mixed training, it has better performance because the conflicts are eliminated, but it cannot utilize the synergy between GUI and embodied data.
$(iii)$ OmniActor achieves better performance than separate training, because it can utilize the synergy between GUI and embodied data while eliminating their conflicts.

\begin{figure}[tb]
\begin{center}
\includegraphics[width=1.0\linewidth]{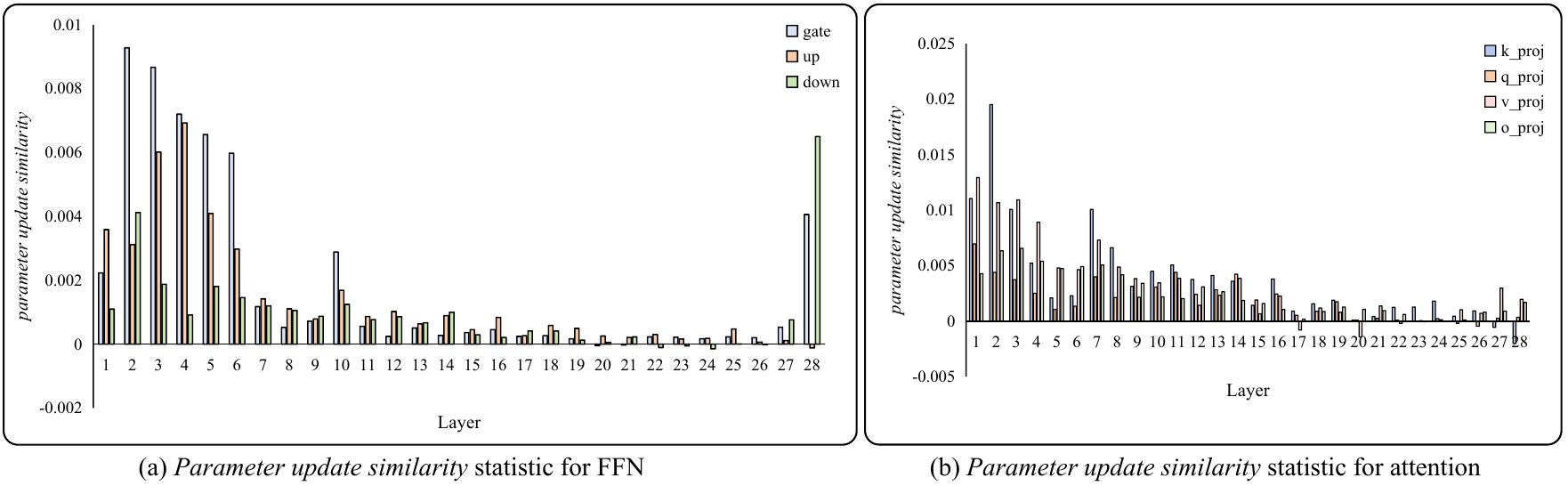}
\end{center}
  \caption{Statistic about \textit{parameter update similarity}.
  Shallow layers have the significantly higher \textit{parameter update similarities} than deep layers, which guides our design for Layer-heterogeneity MoE. 
  Specifically, parameters in shallow layers are shared while parameters in deep layers are separated into two groups: one for GUI tasks and the other for embodied tasks.
  }
\label{statistic1}
\end{figure}

\textbf{Statistic about parameter optimization directions}: We analyze the parameter optimization directions for GUI and embodied tasks.
First, \textit{which parameters can be shared and which should be separated?} We think that the parameters can be shared when the parameter update directions from GUI and embodied data are consistent; otherwise, they should be separated.

\textit{How to find shareable parameters?} We propose a novel metric \textit{parameter update similarity}, estimating the consistency between GUI task and embodied tasks across different parameters.
For example, regarding the gate proj in the first FFN, we calculate the parameter difference before and after training with GUI data, denoted as $d_{gui}$.
We also calculate the parameter difference before and after training with embodied data, denoted as $d_{robot}$.
Then, we compute the cosine similarity between $d_{gui}$ and $d_{robot}$ as the \textit{parameter update similarity} of the gate proj. 
When the \textit{parameter update similarity} is high, the parameter should be shared; otherwise, the parameter should be separated.
We conduct the analysis of FFNs (gate proj, up proj, and down proj) across different layers in Figure~\ref{statistic1} (a); we conduct the analysis of attention (k proj, q proj, v proj, and o proj) in Figure~\ref{statistic1} (b).
Based on the above statistics, we find that shallow-layer parameters should be shared, while deep-layer parameters should be separated.
Based on the above statistic, we empirically set the layer depth threshold $K$, which distinguishes shared and separated layers, to 8.

\begin{table*}[!t]
\small
\setlength\tabcolsep{1.5mm}
\caption{\textbf{Study about different base MLLMs.}
We study different base MLLMs, for verifying the generalization of the proposed strategy. 
We report the success rate on each task.}
\label{tab:study2}
\centering
\begin{tabular}{l|c|ccc|c} 
\toprule
\multirow{2}{*}{Base MLLMs} & \textbf{Embodied Tasks} & \multicolumn{3}{c|}{\textbf{GUI Tasks}} & \multirow{2}{*}{Avg} \\
& LIBERO-90 & AndroidControl-Low & AndroidControl-High & GUI Odyssey & \\
\midrule
Qwen2-VL 7B & 69.5 & 86.4 & 77.5 & 66.0 & 74.9 \\
Qwen2.5-VL 7B & 65.2 & 87.9 & 79.5 & 81.0 & 78.4 \\
\bottomrule
\end{tabular}
\end{table*}

\textbf{Generalization verification}: 
We analyze the impact of training agents on different base MLLMs. 
In detail, we modify the base MLLM to Qwen2.5-VL 7B~\citep{bai2025qwen2}, adjust the training data to match the format compatible with Qwen2.5-VL 7B, and migrate the Layer-heterogeneity MoE to Qwen2.5-VL. 
The experimental results are shown in the Table~\ref{tab:study2}.
When training the generalist agent on the more powerful Qwen2.5-VL 7B, the average success rate increases by 3.5\% compared to training on Qwen2-VL 7B. 
The average success rate on embodied tasks decreases, and the average success rate on GUI Odyssey rises significantly.
We guess that this is because Qwen2.5-VL 7B is more suitable to handle complex cross-application scenarios than Qwen2-VL 7B.

\begin{figure}[tb]
\begin{center}
\includegraphics[width=1.0\linewidth]{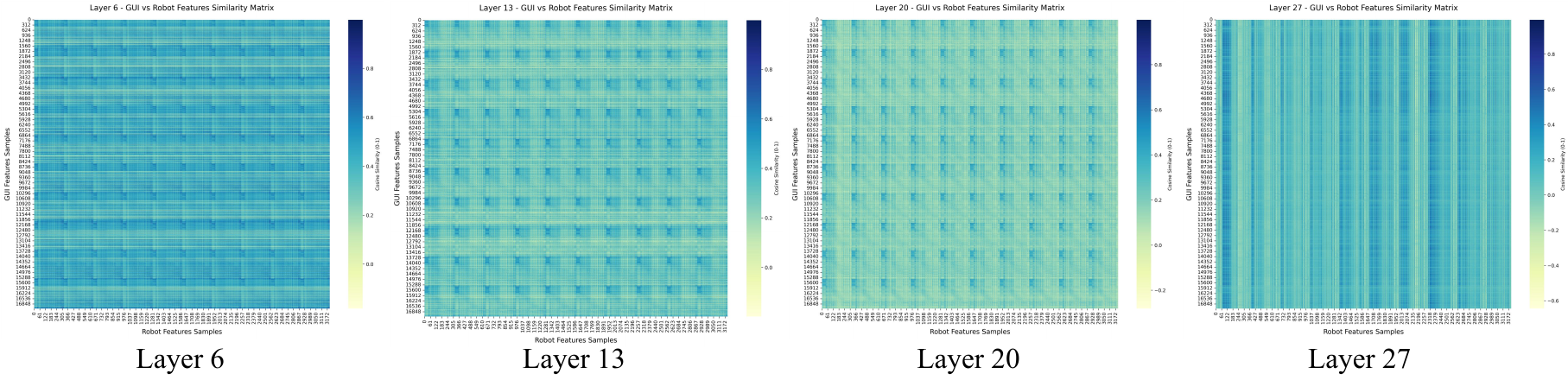}
\end{center}
  \caption{Visualization about synergy and conflict.
  We analyze the relationships between GUI and embodied features in different layers.
  Darker colors indicate higher feature similarity, while lighter colors indicate lower feature similarity.
  }
\label{statistic2}
\end{figure}

\textbf{Visualization about synergy and conflict}:
We conduct a more in-depth analysis of shallow and deep features. 
Specifically, we randomly select a subset of GUI data and use OmniActor-GUI to extract intermediate-layer features $f_{\text{gui}} \in R^{N_{\text{gui}} \times C}$, where $N_{\text{gui}}$ indicates GUI sample count. 
Similarly, we randomly select a subset of embodied data and use OmniActor-EA to extract intermediate-layer features $f_{\text{robot}} \in R^{N_{\text{robot}} \times C}$, where $N_{\text{robot}}$ indicates GUI sample count. 
We then calculate the similarity matrix between $f_{\text{gui}}$ and $f_{\text{robot}}$. 
We focus our analysis on the layers $\{6,13,20,27\}$.
In other words, we generate four similarity matrices. 
By visualizing these similarity matrices in Figure~\ref{statistic2}, we observe that the similarity of shallow features is significantly higher than that of deep features. 
The mean similarity is 0.4206, 0.3018, 0.2000, and 0.1134 for layer 6, 13, 20, and 27, respectively.
This phenomenon further indicates that GUI and embodied tasks exhibit synergy at the shallow layers while showing conflict at the deep layers.

%% file: sec/5_conclusion.tex
\section{Conclusion}
Our research aims to construct a high-performance generalist agent from both data and structural perspectives. Data scaling is particularly crucial for agent performance; therefore, we unify the action spaces of GUI and embodied tasks, and collect a large amount of GUI and embodied data for training the agent. Furthermore, we identify the conflicts and synergies between GUI and embodied tasks. By separating deep-layer parameters and sharing shallow-layer parameters, we leverage the synergies between tasks and eliminate conflicts. Extensive experiments demonstrate the effectiveness of the generalist agent OmniActor across different scenarios. Especially in GUI scenarios, OmniActor even outperforms many of the latest state-of-the-art GUI agents.

One limitation of this work is that the discussed embodied scenarios may be insufficient compared to GUI scenarios. 
In the future, we will supplement more data on embodied scenarios (\textit{e.g.}, CALVIN) and conduct larger-scale experiments.